\documentclass{article}

\usepackage{spconf}
\usepackage[cmex10]{amsmath}
\usepackage{amsthm}
\usepackage{amssymb}
\usepackage{mathrsfs}
\usepackage{graphicx}
\usepackage{float}
\usepackage{array}
\usepackage{epstopdf}
\usepackage{multirow}

\usepackage{amsmath,amssymb}
\usepackage{graphicx}
\usepackage{subfig}
\usepackage{float}

\title{e\MakeLowercase{fficient} s\MakeLowercase{uper} r\MakeLowercase{esolution} f\MakeLowercase{or} l\MakeLowercase{arge}-s\MakeLowercase{cale} i\MakeLowercase{mages} u\MakeLowercase{sing} a\MakeLowercase{ttentional} gan}

\name{Harsh Nilesh Pathak$^*$, Xinxin Li$^{\dagger}$, Shervin Minaee$^{\dagger}$, and Brooke Cowan$^{\dagger}$}
\address{$^{\dagger}$Expedia Group,
Bellevue, WA, USA\\
$^*$Worcester Polytechnic Institute, Worcester, MA, USA \\}

\allowdisplaybreaks[1]	
\begin{document}
%
\maketitle
\begin{abstract}
Single Image Super Resolution (SISR) is a well-researched problem with broad commercial relevance. However, most of the SISR literature focuses on small-size images under 500px, whereas business needs can mandate the generation of very high resolution images. At Expedia Group, we were tasked with generating images of at least 2000px for display on the website --- four times greater than the sizes typically reported in the literature. This requirement poses a challenge that state-of-the-art models, validated on small images, have not been proven to handle. In this paper, we investigate solutions to the problem of generating high-quality images for large-scale\footnote{Though elsewhere in the literature \textit{large-scale} is used to refer to the size of the data set, here we use the term to refer to images with a large number of pixels.} super resolution in a commercial setting. We find that training a generative adversarial network (GAN) with attention from scratch using a large-scale lodging image data set generates images with high PSNR and SSIM scores. We describe a novel attentional SISR model for large-scale images, A-SRGAN, that uses a flexible self attention layer to enable processing of large-scale images. We also describe a distributed algorithm which speeds up training by around a factor of five.
\end{abstract}

\section{Introduction}

\begin{figure}[t]
\begin{center}
\begin{tabular}{ccc}
   \hspace{-0.1cm} {4$\times$ SRGAN \cite{DBLP:journals/corr/LedigTHCATTWS16}} & \hspace{-0.1cm} {4$\times$ A-SRGAN} & \hspace{-0.15cm} {Original} \\
    {} & {(Proposed)} & {} \\
\includegraphics[width=0.3\linewidth]{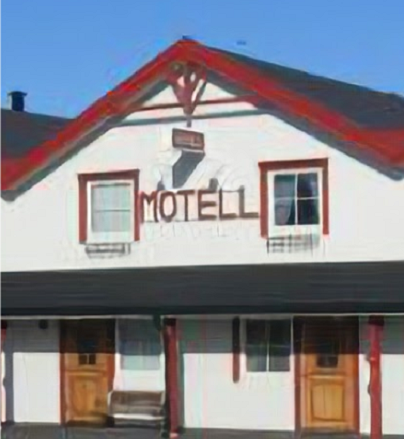} & \hspace{-0.2cm} \includegraphics[width=0.3\linewidth]{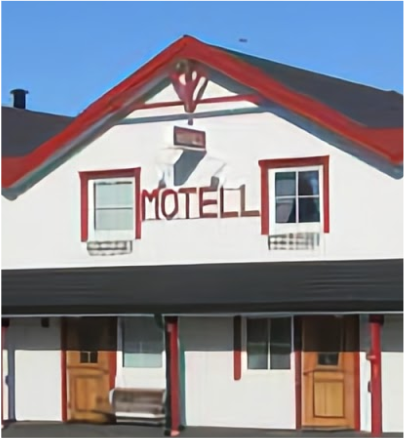} & \hspace{-0.2cm}
\includegraphics[width=0.3\linewidth]{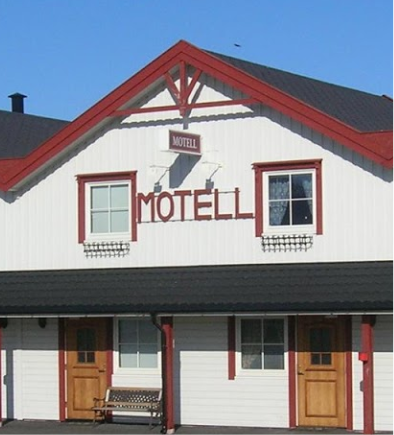}
\\
\includegraphics[width=0.3\linewidth]{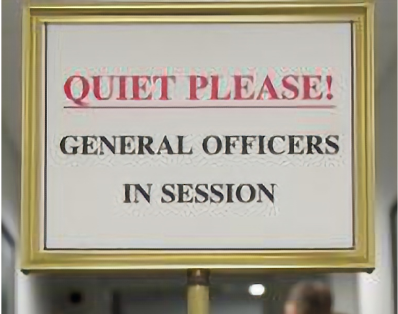} & \hspace{-0.2cm} \includegraphics[width=0.3\linewidth]{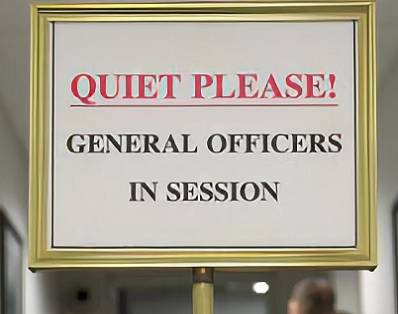} & \hspace{-0.2cm}
\includegraphics[width=0.3\linewidth]{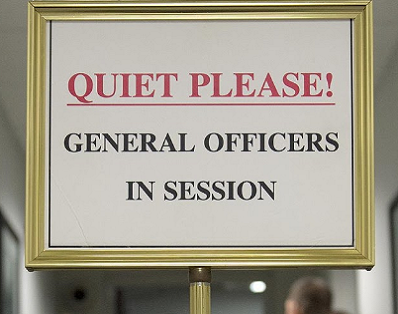} \\
\end{tabular}
\end{center}
\caption{Super-resolution result of our proposed algorithm, compared with SR-GAN model \cite{DBLP:journals/corr/LedigTHCATTWS16}. As we can see the proposed attentional-GAN model performs better in upsampling regions around text and fine details, whereas SRGAN  results are blurry in those regions.}
\vspace{-3mm}
\end{figure}

The task of producing a high-resolution (HR) image given its low-resolution (LR) image counterpart is known as \textit{super resolution} (SR)\footnote{Throughout this paper, we borrow the following notations from \cite{DBLP:journals/corr/LedigTHCATTWS16}: \begin{math}I^{LR}\end{math} (\textit{low-resolution image}), \begin{math}I^{HR}\end{math} (\textit{high-resolution image}), and \begin{math}I^{SR}\end{math} (\textit{super-resolution image}).}. Super resolution is an under-determined inverse problem and thus quite challenging, since a given LR pixel may lead to multiple solutions based on variant texture details in the corresponding HR image \cite{DBLP:journals/corr/LedigTHCATTWS16},\cite{DBLP:journals/corr/DongLHT15}. Most contemporary work on super resolution focuses on Single Image Super Resolution (SISR), which can handle input images of arbitrary size. This is because at test time, the system processes a single image as opposed to a batch of images \cite{DBLP:journals/corr/LedigTHCATTWS16}, \cite{DBLP:journals/corr/ShiCHTABRW16}. SISR is a well-researched problem with broad commercial application to many image and video analysis tasks, such as video surveillance, image/video streaming \cite{DBLP:journals/corr/ShiCHTABRW16} and image medical investigation \cite{7317595}. However, to date, published work on SISR has been validated on small-size images of fewer than 500px (e.g., SET5 \cite{set5}, SET14 \cite{set14}, BSD100 \cite{bsd100}). In contrast, business needs may mandate the generation of very high resolution images. We focus here on the problem of generating large-scale SR images in a commercial setting.

At Expedia Group, the business requires images of at least 2000px for display on the website --- four times greater than the sizes typically reported in the literature. This requirement presents an interesting research challenge since, to the best of our knowledge, state-of-the-art models for SISR, e.g., SRGAN \cite{DBLP:journals/corr/LedigTHCATTWS16}, SRCNN \cite{DBLP:journals/corr/DongLHT15}, and ESPCNN, \cite{DBLP:journals/corr/ShiCHTABRW16}, have not been tested in large-scale resolution spaces. Large-scale images are more susceptible to the object consistency problem \cite{sagan} due to the larger number of pixels and long-term dependencies across different regions of the image. Furthermore, large-scale images incur greater computational cost in both training and inference. Our goal is to develop a model with high visual quality output that performs efficiently on larger-scale images during training and testing. From the quality standpoint, we measure success in terms of two standard SR metrics, Peak Signal-to-Noise Ratio (PSNR) and Structural SIMilarity Index (SSIM) \cite{Wang04imagequality}, setting targets of PSNR $>$ 25 and SSIM $>$ 0.75. 

To solve this business problem, we investigate several approaches. First, we simply apply a pre-trained state-of-the-art SR model, trained on large-scale images, to our Expedia Group lodging dataset. Because this does not satisfy our PSNR and SSIM target values, we next try fine-tuning the weights of the pre-trained model using an 11K-subset Expedia Group lodging images. While results are satisfactory according to PSNR and SSIM, upon inspection we find that output from the model contains undesirable artifacts such as ringing near edges and blurriness, which we attribute to object inconsistency in these larger-scale images. In an attempt to overcome these problems, we train two models end-to-end on the full set of 21,997 Expedia Group lodging images. The first is SRGAN \cite{sagan}, a model known for encouraging perceptually-rich images and addressing deficiencies in texture quality exhibited by other models. The second is a novel model called an Attentional Super Resolution Generative Adversarial Network (A-SRGAN). A-SRGAN inserts a Flexible Self Attention layer, or FSA, in the SRGAN architecture, which in theory should provide additional support for texture quality and object consistency in our large-scale resolution space. The FSA layer borrows the self attention layer from SAGAN \cite{sagan}, and wraps it with a max-pooling layer to reduce the size of the feature maps and enable large-size images to fit in memory. We report promising results using A-SRGAN to super-resolve large-scale Expedia Group lodging data.

Finally, in order to run the GAN training regimen efficiently in end-to-end scenarios, we describe a distributed algorithm which significantly speeds up the training process, an important consideration in a business context where both cost and time savings are highly valued. All operations (i.e., matrix multiplications) are computationally more expensive with large-scale data and larger feature maps, thus motivating the need for learning in a distributed framework. However, multi-GPU training with GANs is not as straightforward as training a very deep convolution network, since in GANs we train two neural networks alternately. We use data parallelism \cite{DBLP:journals/corr/YadanATR13} across multiple GPUs to achieve scalability and computational efficiency and demonstrate a boost in training time of about a factor of five.

The contributions of this paper are as follows:
\begin{itemize}
\item We introduce a novel super-resolution algorithm based on attentional GAN for large-scale data.
\item We validate the use of the state-of-the-art SRGAN \cite{DBLP:journals/corr/LedigTHCATTWS16} model on large-scale data.
\item We demonstrate the importance of selecting an appropriate crop size when processing large-scale data.
\item We describe a distributed algorithm for training GANs which speeds up the training process by roughly a factor of five.
\end{itemize}

The remainder of this paper is structured as follows: Section 2 covers related work. Section 3 describes our methodology and approach, including a detailed description of our novel Attentional-SRGAN (A-SRGAN) model, an overview of some of the challenges specific to large-scale data, and an explanation of our distributed, multi-GPU GAN training algorithm. In Section 4, we discuss our experimental results. Finally, Section 5 concludes the paper.

\section{Related Work}
Early approaches to super resolution (e.g., \cite{IRANI1991231},\cite{6751227}) used simple reconstruction-based techniques and transposed convolution networks to recover HR images. This frequently introduced ringing artifacts around the sharp structures \cite{6751227} due to inaccurate kernels. Interpolation methods, such as bicubic and nearest-neighbor, are quite commonly used and can achieve an acceptable trade-off between visual quality and computational efficiency. However, such methods have many limitations due to the use of a predefined kernel. Despite the recent development of adaptive kernels for interpolation as a remedial measure \cite{4697303},\cite{DBLP:journals/corr/abs-1210-3404},\cite{6619119}, this approach has yet to achieve photo-realistic visual quality. 

More recently, deep learning approaches have produced favorable results in many computer vision tasks including SR. There are many interesting deep learning frameworks for SR, but here we focus on three models of particular relevance to our work, SRGAN \cite{DBLP:journals/corr/LedigTHCATTWS16}, SAGAN \cite{sagan}, and SNGAN \cite{snorm}. 

\subsection{Super-Resolution GAN (SRGAN)}

Super-Resolution through Generative Adversarial Networks (SRGAN)   \cite{DBLP:journals/corr/LedigTHCATTWS16} is a generative model that uses GANs \cite{gan} to produce an estimate of the HR image given the LR image. The SRGAN architecture involves first training a ResNet-based model \cite{DBLP:journals/corr/HeZRS15} called SRResNet ---
a deep residual network with convolutions and a pixel shuffle layer \cite{DBLP:journals/corr/ShiCHTABRW16}) that minimizes the \textit{content loss} between $I^{SR}$ and $I^{HR}$. 
SRGAN uses SRResNet as the generator and takes an adversarial approach to estimate the probability distribution of the target dataset.

SRGAN encourages perceptually-rich images and directly addresses deficiencies in texture quality exhibited by earlier models (SRCNN \cite{DBLP:journals/corr/DongLHT15}, ESPCNN \cite{DBLP:journals/corr/ShiCHTABRW16}, and SRResNet itself). The \textit{perceptual loss function}, the weighted sum of a \textit{content loss} and an \textit{adversarial loss}, is a crucial theoretical contribution of SRGAN used to recover fine texture content. 
There are two versions of the SRGAN content loss: (1) a straightforward pixel-wise MSE loss between the $I^{SR}$  and $I^{LR}$, similar to ESPCNN, and (2) VGG loss, which is the Euclidean distance between 
feature maps of $I^{SR}$ and $I^{HR}$ passed through the VGG19 network. The adversarial loss enhances the texture quality of the image, as a GAN drives the distribution of the generated image $I^{SR}$ to the true distribution $I^{LR}$, which is in the natural image manifold, thus enhancing the visual appearance of the image. 

The SRGAN authors report state-of-the-art results according to the mean opinion score (MOS) of human judges, a strong indication of visual quality. Interestingly, though both PSNR and SSIM decrease between training the ResNet and the GAN parts of SRGAN, MOS increases. SRGAN's strength in visual and texture quality was a strong influence in our decision to use it for our work with large-scale data, where visual quality is of paramount importance. However, the model as described in the paper \cite{DBLP:journals/corr/LedigTHCATTWS16} is trained with ImageNet data \cite{imagenet_cvpr09}, which contains mostly small-scale with resolutions of between 200--500px. Moreover, SRGAN was tested on SET5 \cite{set5}, SET14 \cite{set14}, and BSD100 \cite{bsd100}, which contain images of no more than 500px. While SRGAN is widely-accepted as a state-of-the-art SR model, to the best of our knowledge its efficacy on large-scale data has been unexplored prior to this work.

\subsection{Self Attention GAN (SAGAN)}
The SAGAN model \cite{sagan} introduces a \textit{self attention layer} for capturing long-term dependencies in a general GAN framework. This layer takes input from the previous convolution layer and outputs feature maps of the same size. The architecture is inspired from the query and key model in \cite{DBLP:journals/corr/VaswaniSPUJGKP17}. The input feature maps are transformed into the spaces of a query $f$ and a key $g$, where $f(x) = W_f x$ and $g(x) = W_g x$. Then the attention map is calculated as follows: 
\begin{equation}\label{eq:atten}
    \beta_{i,j} = \frac{exp(s_{ij})}{\sum_{i=1}^N exp(s_{ij})} 
\end{equation}
where \begin{math} s_{ij} = f(x)^{T}g(x)\end{math} and $\beta_{i,j}$ implies the extent to which the model attends to the $i^{th}$ location when synthesizing the $j^{th}$
region. This attention map ($\beta_{i,j}$) is included in the feature maps, and a weighted skip connection is established to make sure that the attention is added on top of the feature maps of previous layer.

Although SAGAN in theory handles the kinds of long-term dependencies frequently exhibited in large-scale data, we find that in practice it is limited to small images and quickly runs out of memory on our data (e.g., when the resolution space is 2000px with 3-channels for RGB-images). Our work extends SAGAN by introducing a Flexible Self Attention layer that handles large-size images by adding a max-pooling layer, as explained in Section IV.

\subsection{Spectral Normalization GAN (SNGAN)}
Weight normalization \cite{wnorm} is a computationally cheap and efficient technique that has proved highly effective in many computer vision models. Spectral normalization is an extension of weight normalization, where the weights of a hidden layer are normalized by the largest singular value of the weight matrix of the same layer while satisfying local 1-Lipschitz constraint \cite{snorm}. Empirically, this technique converges faster and better, particularly in GANs \cite{weighthnorm}, as spectral normalization has been shown to help with the mode collapse problem that GANs often face \cite{snorm}. The Spectral Normalization GAN (SNGAN) \cite{snorm} uses spectral normalization by constraining the Lipschitz constant of the discriminator network. Note that conditioning the weights of the generator network in addition to the discriminator has also proven successful \cite{sagan}. Hence, borrowing from these previous works, we use spectral normalization in all of our convolutional and dense layers in both the generator and discriminator networks of A-SRGAN. 

This section describes how we address some of the key challenges of generating high quality images in a large-scale resolution space. First, we describe a novel super resolution GAN model with attention. Next, we discuss the importance of selecting an appropriate crop size for our large-scale data. Finally, we explain how we leverage distributed, multi-GPU computation for efficient GAN training.

\section{Attentional Super Resolution GAN (A-SRGAN)}
Our A-SRGAN model (shown in Fig 3) is an extension of SRGAN intended to provide additional support for high visual quality in large-scale images using attention. We introduce two modifications to SRGAN: (1) we include an attention layer called Flexible Self Attention (FSA) to account for long-term dependencies and potentially correct the effects of object inconsistency that we see when fine-tuning, e.g., ringing at the edges of walls and tables; (2) we introduce layer-wise spectral normalization for every convolutional and dense layer to condition the weights of the network.

\textbf{Flexible Self Attention layer}:
The idea of including an attentional component in the SR task is to capture long-term multilevel dependencies across image regions that are far apart and unseen by kernels \cite{sagan}. 
However, as explained in Section II, the amount of memory required to store the correlation matrix (i.e., attention map) of SAGAN's self attention layer is prohibitively large for large-scale images. For instance, the flattened correlation matrix \begin{math}s_{ij}\end{math} (Equation~\ref{eq:atten}) for an input image of size 500x500px is 250Kx250Kpx, which is quite costly to store in-memory. 

\begin{figure*}
        \begin{center}
        \includegraphics[scale=0.4]{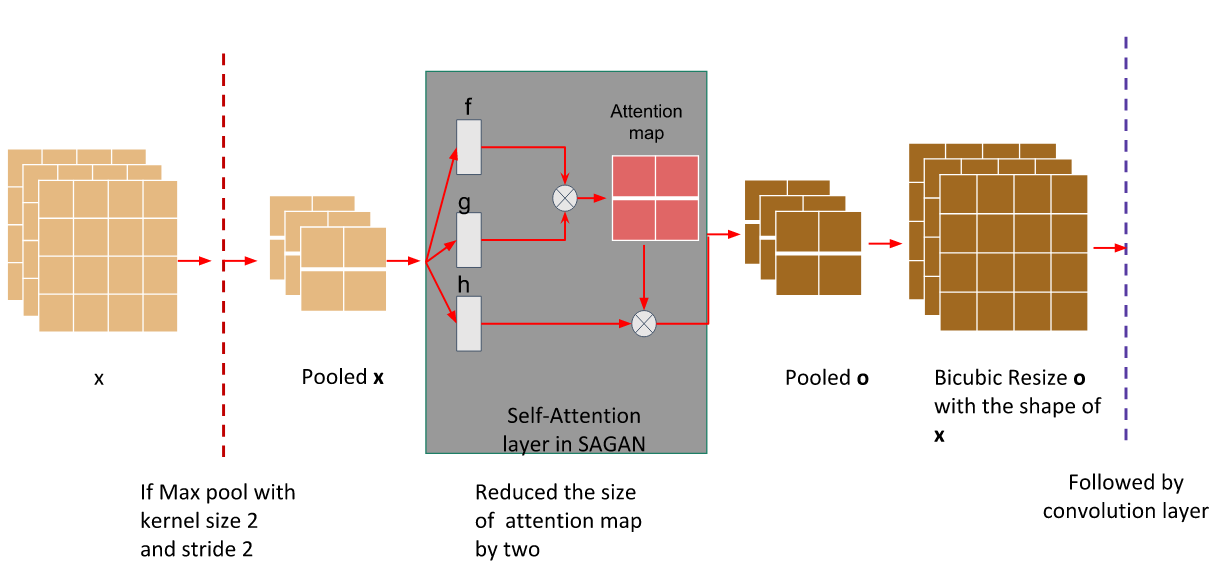}
        \caption{Flexible Self Attention layer (FSA). The input $x$ is a set of feature maps from the previous convolutional layer, and the output $o$ is resized with the shape of $x$.  
         The self attention layer in SAGAN \cite{sagan} has three functions $f$, $g$, $h$, which  have learnable weights as  $1\times1$ convolutions. The attention map in this figure refers to the output of softmax being applied to the raw attention map. We wrap this layer with a pooling layer and a deconvolution, forming a flexible self-attention layer that can be inserted between any two convolutional layers.}
        \label{fig:flexatten}
        \end{center}
\end{figure*}

Our FSA layer adds attention to the model without exploding memory for large-scale images. We wrap the SAGAN self attention layer with max-pooling and then resize the image to match the shape of the input, as shown in Figure~\ref{fig:flexatten}. Since the input and output feature maps are of the same size, the FSA can be inserted between any two convolutional layers. The wrapping reduces the size of the attention map, enabling us to perform attention on large size images. The size of the attention map (\begin{math}s_{ij}\end{math} in equation \ref{eq:atten}) is reduced by the square of the pool size $p$.

We include the FSA layer in both the generator and discriminator, and we refer to the generator network when trained with attention as A-SRResNet. We employ these FSA layers in testing to handle the large-size input images; during training, because the input image size is small due to cropping, we use the SAGAN \cite{sagan} self attention layer. However, note that the FSA layer does not add any new variables and can be used in either training or testing as needed.

\begin{figure*}
        \begin{center}
        \includegraphics[scale=0.2]{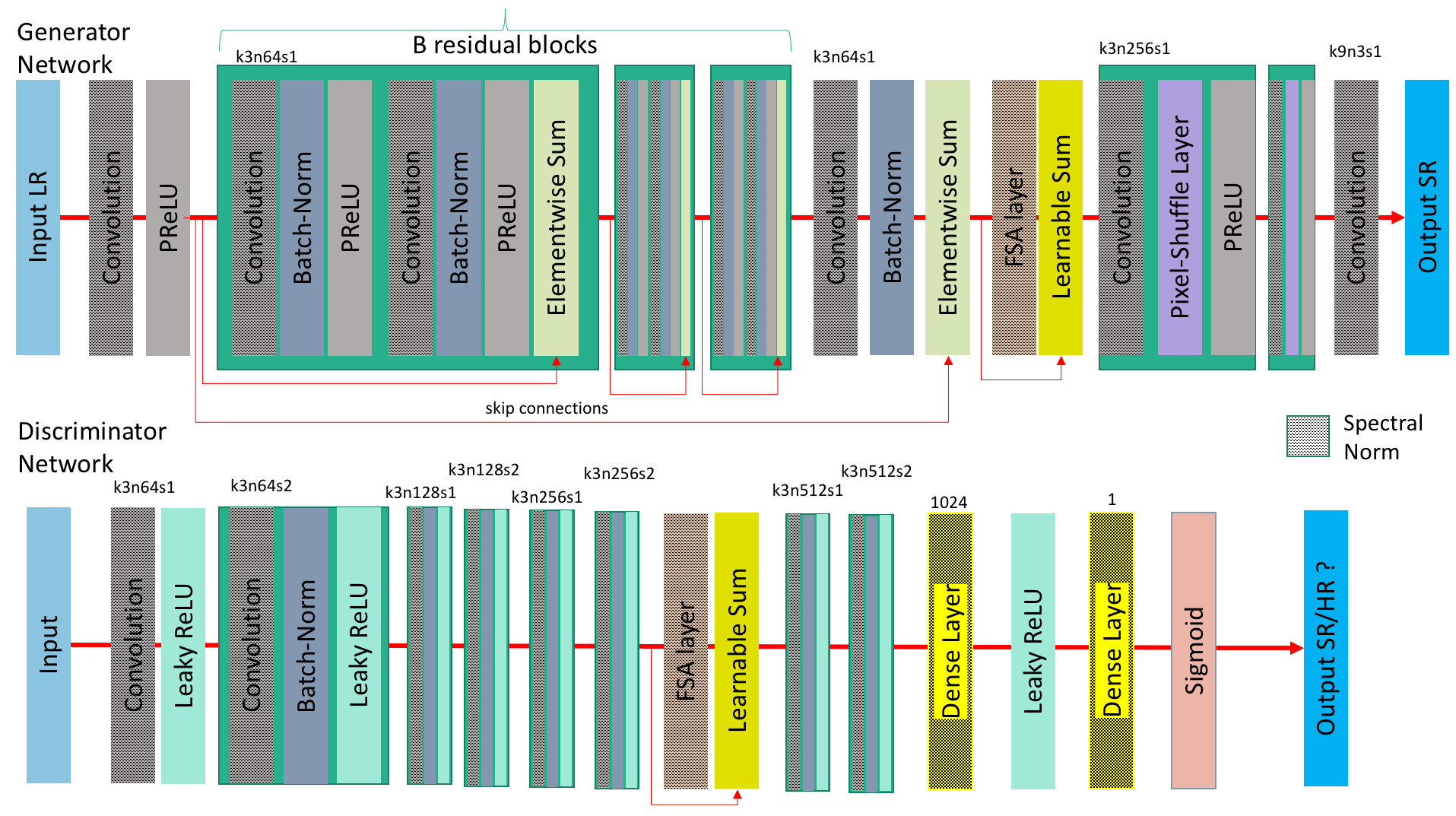}
        \caption{The A-SRGAN architecture extends SRGAN \cite{DBLP:journals/corr/LedigTHCATTWS16} with an FSA layer inspired by SAGAN \cite{sagan}. Note that the \textit{Learnable Sum} operation refers to the weighted skip connection from SAGAN \cite{sagan}. In each layer, the weights are normalized using spectral normalization. The generator and discriminator networks of A-SRGAN are shown with their corresponding kernel size ($k$), number of feature maps ($n$) and stride ($s$). The generator network is referred as A-SRResNet, trained with MSE loss.}
        \label{fig:aSRGAN}
        \end{center}
\end{figure*}

\subsection{Crop Size Selection}

Most state-of-the-art models for SISR, including SRGAN \cite{DBLP:journals/corr/LedigTHCATTWS16}, SRCNN \cite{DBLP:journals/corr/DongLHT15}, and ESPCNN, \cite{DBLP:journals/corr/ShiCHTABRW16} employ a fixed-size, random crop of the LR image during training. Typically, the crop size used is 24px. However, it's not obvious that this size works well for all resolution spaces, and in particular for large-scale images. We hypothesize that selecting an appropriate crop size is critical for achieving high visual quality SR images in large-scale data and therefore treat it as an optimizable hyperparameter.

\begin{figure}[h]
        \begin{center}
        \subfloat[]{\includegraphics[ scale=0.40]{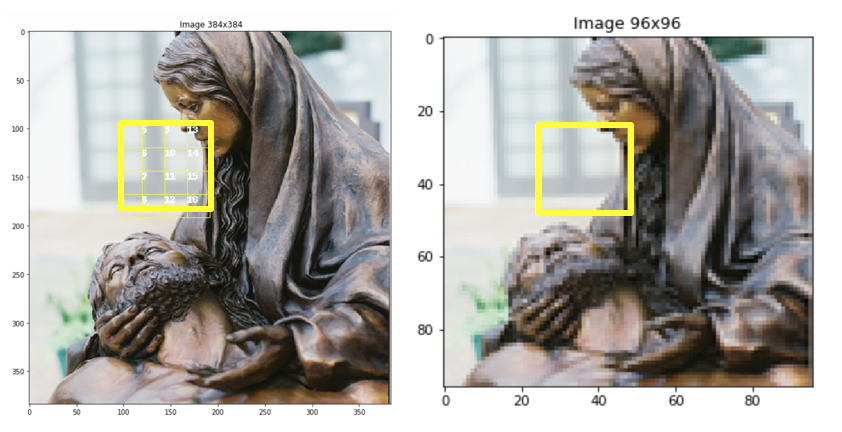}
        }
        \end{center}
       
        \begin{center}
        \subfloat[]{\includegraphics[ scale=0.40]{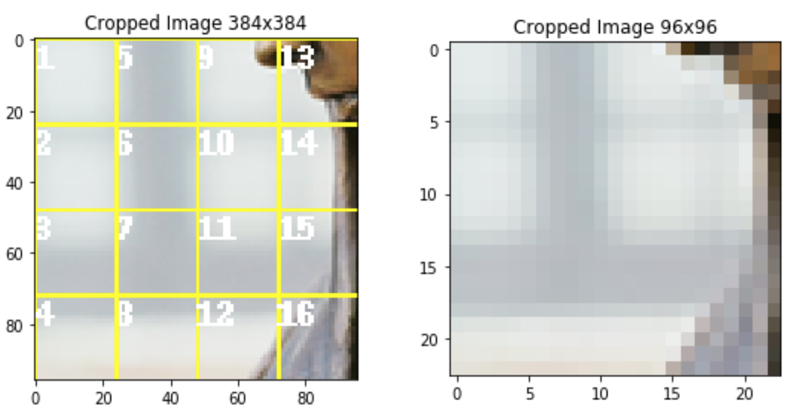}}
        \end{center}
        \caption{Crop Size Selection (a) The image on the left is 384x384px, 4 times larger than the same image on the right (96x96px). Both images have been resized using bicubic interpolation. In (b), we see that a crop size 24px captures far more information and long-term dependencies in the smaller resolution image. Though the sub-image (shown in yellow) in both cases is equivalent, the left sub-image requires 16 crops of size 24px to capture similar information about long-term dependencies when compared to the single crop on the right. Many of the 24px crops in the left image are in fact redundant and individually capture similar pixel values.}
        \label{fig:Crop}
\end{figure}

Figure~\ref{fig:Crop} provides some intuition for the importance of choosing a crop size that is well-suited to the resolution space. The figure contains two images that have been resized using bicubic interpolation. We observe that when the training input to the model is from a small-scale resolution space  (e.g., 96x96px), a standard 24px crop size captures a significant amount of information and long-term dependencies. However, if we apply the same crop size to the image in a larger-scale resolution space (384x384px), we need 16 crops of size 24px to get similar information about long-term dependencies. While redundancy and homogeneity in individual crops may help model texture quality, the literature suggests that feature maps with greater information (entropy) perform better for various tasks ( \cite{Erhan_2014_CVPR}). We investigate the effect of crop size on visual quality for the super resolution task in Section 3.

\subsection{Distributed GAN Training}
\begin{figure*}
        \begin{center}
        \includegraphics[scale=0.5]{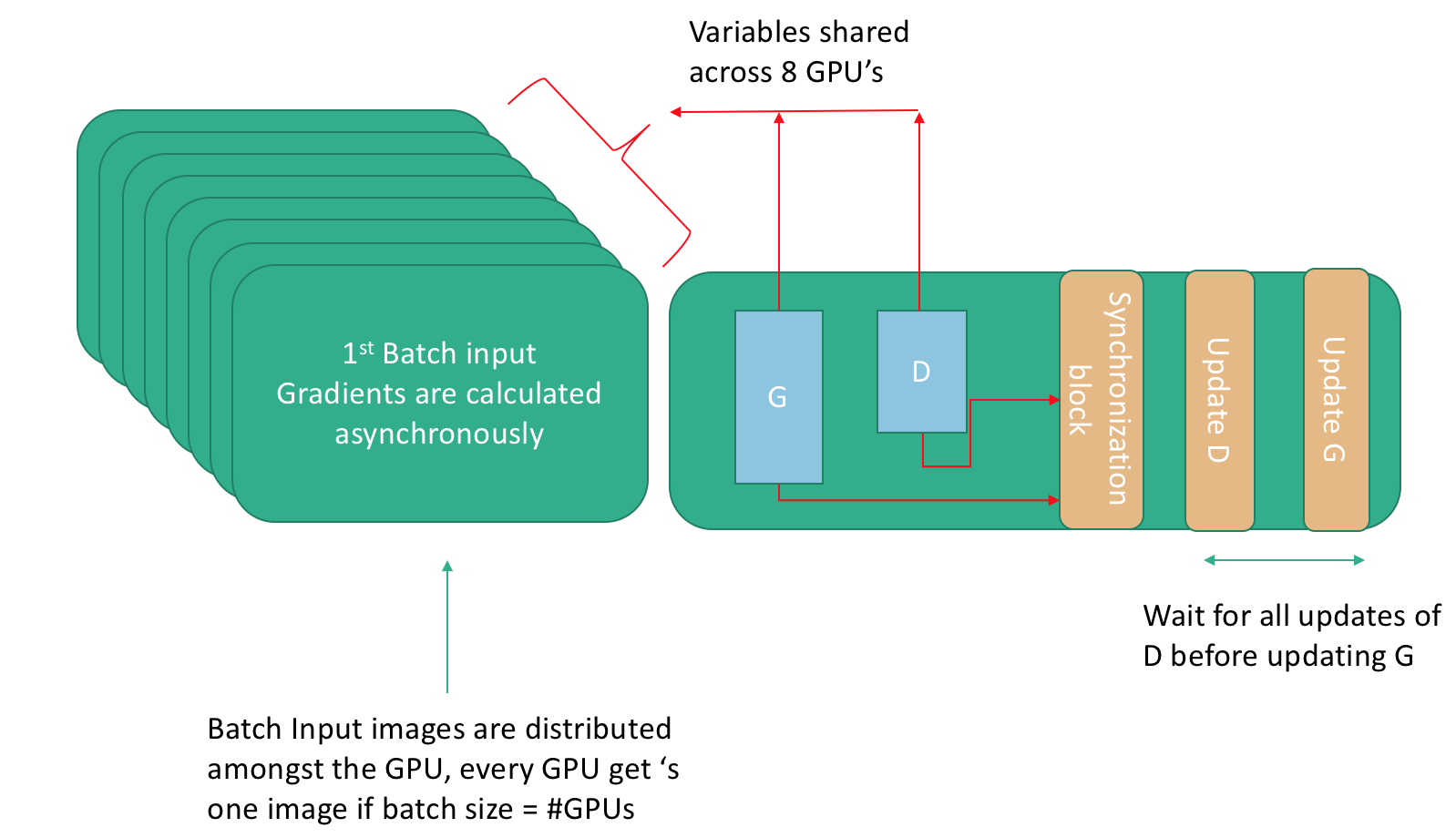}
        \end{center}
        \caption{Distributed GAN flow. The green blocks on the left represent GPU:0-7, and GPU:0 is shown on the right. $G$ refers to the generator network and $D$ to the discriminator network of the GAN. The variables of both $G$ and $D$ are stored in GPU:0 and shared across all GPUs. When a batch of images is being processed, the gradients are calculated independently over all GPUs and then updated synchronously in GPU:0.}  
        \label{fig:DSRGAN}
\end{figure*}

In order to improve the efficiency of training GANs, we develop a method for distributing the computation over multiple GPUs. Figure~\ref{fig:DSRGAN} shows how we achieve data parallelism with GANs. In order to distribute the training, we first perform the updates on the discriminator ($D$), and then we update the generator ($G$) using a synchronization block. All other gradient calculations are performed asynchronously on each GPU with batch inputs divided equally over all GPUs.

\section{Experiments}
\subsection{Experimental setup}
We use data from the Expedia Group lodging image database for training and testing the models. This data set contains more than 15M distinct images of varying resolution. In accordance with business requirements, we define our HR target resolution space to be 2000px. Accordingly, for our HR images, we select images from the data set of around size 2000x1340px, with some small variations (e.g., 2000x1333px, 2000x1350px, etc.). The LR images are around 500x350px (4x smaller), with corresponding variations. After filtering the original data set for images of the appropriate size, we designate 21,997 images for training, and 995 for test. We report PSNR (Peak Signal-to-Noise Ratio) and SSIM (Structural SIMilarity Index) \cite{Wang04imagequality} on this test set across all experiments. For pre-processing, we normalize our LR images to [0,1] and HR images to [-1,1].  

Legal restrictions on the Expedia Group images prevented us from publishing them, so we created an auxiliary data set consisting of 47 similar images in the public domain that we use for illustrating model output (see Figure 6). This Google-47 data set was downloaded via the $google-images-download$ API \cite{gid} with specific arguments (size: "\textgreater 2MP", user-rights: "labeled-for-reuse-with-modifications", keywords: "room", "motel" and "hotel" and aspect-ratio: "wide"). 
Before using these images, we resized them 
to 2000x1400px for consistency with the resolution space of the Expedia Group data, and induced the corresponding LR images by downsampling to 500x350px.  


Given the computational demands of the SRGAN and A-SRGAN models, we elected to use Amazon EC2 instances for all our experiments. More specifically, we used Deep Learning Ubuntu instances with 32 CPUs, 488GB memory, and 8 NVIDIA K80 GPUs. Furthermore, we created an anaconda environment with Python 2.7 and tensorflow-gpu 1.8. We find that our distributed GAN training framework significantly speeds up training in the end-to-end framework, reducing execution time by roughly a factor of five (Table~\ref{table:execution-time}). We leverage this distributed framework throughout all experiments described in this section.

\begin{table}[h]
\begin{center}
\begin{tabular}{|c|c|c|c|c|}  
\hline
Model & GPUs & Execution time & PSNR & SSIM  \\ 
\hline
Model & GPUs & Execution time & PSNR & SSIM  \\ 
\hline
SRGAN\_{MSE} &  1 & 17 hrs & 14.60 & 0.679\\
\hline
SRGAN\_{MSE} & 8 & 3.8 hrs & 14.56 & 0.679 \\
\hline
\end{tabular}
\caption{Distributing GAN training over 8 GPUs speeds up execution time by roughly a factor of 5. The models are trained for $3\times 10^{4}$ steps with a crop size of 24px using 11K large-scale Expedia Group lodging images.}
\label{table:execution-time}
\end{center}
\end{table}

\subsection{Baseline model: Pretrained SRGAN}
As a baseline, we apply a pre-trained SRGAN classifier \cite{bworld} to our large-scale Expedia Group lodging test set. The classifier is trained with the RAISE dataset \cite{Dang-Nguyen:2015:RRI:2713168.2713194} which comprises HR images from a large-scale resolution space (8,156 images ranging from 2500 to 5000px). Thus, we have reason to believe it is more compatible with our target resolution space than if it had been trained on smaller scale images in other publicly-available data sets such as ImageNet. The pre-trained classifier was trained in \begin{math}10^6 SRResNet + 3 \times10^5 SRGAN_{MSE} + 2 \times 10^5 SRGAN_{VGG}\end{math} steps. Results on our test set are shown in the first row of Table~\ref{pre-fine-results}, with PSNR of 22.36, and SSIM of 0.608. 

\subsection{Fine-tuning}
Since PSNR and SSIM are both too low using the pre-trained classifier to satisfy our business requirements (PSNR $>$ 25, SSIM $>$ 0.75), we explore transfer learning by fine-tuning the weights of the pre-trained model. Again, we expect that using a base model trained on large-scale data will be advantageous in a transfer learning setting with similar target large-scale resolution space. We use 11K images chosen at random from our training set to fine-tune the weights in an additional $3\times 10^4$ backpropagation steps for SRGAN$_{MSE}$.  Then, the SRGAN$_{VGG}$ weights are initialized using the weights from the preceding step, and the fine-tuning resumes for an additional $3\times 10^4$ steps. For the VGG loss, we used the factor of 0.0061 as provided by the SRGAN authors \cite{DBLP:journals/corr/LedigTHCATTWS16}. We use a learning of \begin{math}10^{-5}\end{math}, and the same number of global steps for the generator and discriminator ($3\times 10^4$). The generator network has 16 residual blocks, identical to SRGAN.

Results for fine-tuning are shown in rows 2 and 3 of Table~\ref{pre-fine-results}. We note that, similar to reported results in the SRGAN literature \cite{DBLP:journals/corr/LedigTHCATTWS16}, the additional weight transfer to $SRGAN_{VGG}$ decreases both PSNR and SSIM. Due to time and resourcing constraints, we were unable to verify whether mean opinion score increases in spite of the degradation in PSNR/SSIM as in the SRGAN paper. However, informal inspection of the results suggests that the VGG step also introduces new random artifacts. It is possible that tuning the VGG scaling factor for our large-scale data might help. We leave exploration of these issues to future work. Given the lower scores and possible introduction of undesirable artifacts, for the remaining experiments we consider only the MSE step when training SRGAN and A-SRGAN.

Although the mean SRGAN$_{MSE}$ PSNR and SSIM scores fall within an acceptable range according to our targets, we observe undesirable artifacts such as ringing in the structural content (e.g., windows, table edges, etc.) of many images (see Figure 6). We hypothesize that these effects are related to the failure of the fine-tuning to adequately model the long-term dependencies of Expedia Group's large-scale images. These results motivate exploration of an end-to-end approach and development of a model better suited to handle object consistency in our data. 

\begin{table*}\label{pre-fine-results}
\begin{center}
\begin{tabular}{|c|c|c|c|c|c|} 
\hline
 Model & Steps & GPUs & Execution time & PSNR & SSIM \\ [0.5ex] 
 \hline\hline
 PreTrained & \begin{math}10^6+ (5 \times 10^5)\end{math}  & 1 & - & 22.36 & 0.608\\ 
 SRGAN$_{MSE}$ & \begin{math}3 \times 10^{4}\end{math} & 8 & 3.8 hrs & \textbf{27.68} & \textbf{0.784} \\
 SRGAN$_{VGG}$ & $3\times 10^{4}$ & 8 & 3.8 hrs & 27.50 & 0.775 \\ [1ex] 
 \hline
\end{tabular}
\caption{
Transferring the weights from a pre-trained model to SRGAN$_{MSE}$ dramatically increases both PSNR and SSIM. Note that an additional weight transfer to SRGAN$_{VGG}$ decreases both PSNR and SSIM. The execution time for each weight-transfer step on eight GPUs, using the distributed GPU training is just under 4 hours. All models use a crop size of 24px, consistent with the pre-trained model. Fine-tuning was performed using 11K Expedia Group large-scale lodging images.}
\label{pre-fine-results}
\end{center}
\end{table*}

\begin{table*}
\begin{center}
    \begin{tabular}{|c|c|c|c|c|}  
    \hline
     Task  & Global steps & Execution time & PSNR & SSIM  \\ [0.5ex] 
    \hline
    SRResNet & \begin{math}10^6\end{math} & 2 days 20.42 hrs & \textbf{26.87} & 0.795 \\
    SRGAN$_{MSE}$ &  \begin{math}10^5\end{math} & 19.45 hrs  & 26.51 & 0.756 \\
     \hline
     \hline    
      A-SRResNet & \begin{math}10^6\end{math} & 3 days 8.68 hrs & 26.60 & \textbf{0.802} \\
      A-SRGAN$_{MSE}$ &  \begin{math}10^5\end{math}  & 23.32 hrs  & 26.44 & 0.738 \\
     \hline
    \end{tabular}
\caption{A-SRResNet generates the highest SSIM value when trained end-to-end on just over 21K large-scale Expedia Group lodging images, while SRResNet performs better in PSNR. Note that the additional GAN training degrades performance in both models. A crop size of 72px is used across all models.}
\label{table:4}
\end{center}
\end{table*}

\subsection{End-to-end training}
In this section, we explore two models in an end-to-end training framework.  Table~\ref{table:4} shows results on our test set after training SRGAN and A-SRGAN from scratch on the full training set of 21,997 large-scale Expedia Group lodging images. When training both models, we randomly initialize the ResNet and then transfer its weights to the GAN, following \cite{DBLP:journals/corr/abs-1710-10196} and \cite{DBLP:journals/corr/LedigTHCATTWS16}.

As discussed in Section III, intuitively, a larger crop size should help capture more long-term dependencies in large-scale data. Table~\ref{table:crop-size} shows empirically that this is indeed the case. We see that a crop size of 96px generates significantly higher PSNR and SSIM scores than a crop-size of 24px on our large-scale test set. When training SRGAN and A-SRGAN from scratch (Table~\ref{table:4}), we choose a crop size of 72px, which shortened training times and did not lead to any memory issues during training. Finally, we note that using the smallest possible pool size $p$ that allows the feature maps to fit in memory at test time is optimal in terms of PSNR (Table~\ref{table:pooling}). Throughout our experiments with A-SRGAN, we set $p=4$ for the pooling in the FSA layers. The learning rate used in Tables~\ref{table:4}, \ref{table:crop-size}, and \ref{table:pooling} is $10^{-4}$.

Notably, the results in Table~\ref{table:4} show that training both SRGAN and A-SRGAN\footnote{We ran experiments with and without spectral norm regularization, and checked the intermediate PSNR/SSIM values of 995 test images. We observed faster convergence with spectral norm, but after 100k steps we got similar PSNR/SSIM values.} improves SSIM compared with fine-tuning (Table~\ref{pre-fine-results}). In an informal visual inspection of the results of all models, we observe that end-to-end training generates SR images with noticeably superior visual quality when compared to fine-tuning (see Figure 6). This leads us to believe that SSIM is a better indicator of visual quality than PSNR, which decreases in the end-to-end context relative to fine-tuning. Additionally, we note that our attentional model achieves the highest SSIM score (0.802) of all trained models, which is an encouraging result suggesting that there is value in further investigation of attention in large-scale images in future work. The successful application of SRGAN to large-scale images is also encouraging, since SRGAN has not been tested on such data in previous work.

\begin{table}[h]
\begin{center}
\begin{tabular}{|c|c|c|c|c|}  
\hline
Model & Crop Size & execution time & PSNR & SSIM  \\ [0.5ex]
 \hline\hline
 A-SRGAN & 24 & 3.8 hrs & 16.92 & 0.719 \\
 \hline \hline
 A-SRGAN & 96 & 4.4 hrs & 23.82 & 0.765 \\[1ex]
 \hline
\end{tabular}
\caption{Here we investigate the impact that crop size has on PSNR, SSIM and execution time. For large-scale data, a 96px crop size leads to markedly better performance in terms of PSNR and SSIM, but also adds overhead in terms of execution time. Both models were trained for $3\times 10^4$ steps on 8 GPUs, using 11K large-scale Expedia Group lodging images.}
\label{table:crop-size}
\end{center}
\end{table}

\begin{table}[h]
\begin{center}
  \begin{tabular}{|c|c|c|c|}  
\hline
 Model  & Pool size (p) & PSNR & SSIM  \\ [0.5ex]
 \hline\hline
 A-SRResNet & 32  & 23.13 & 0.761 \\
 A-SRResNet & 16  & 23.35 & 0.764 \\
 A-SRResNet & 8  & 23.64 & 0.766 \\
 A-SRResNet & 4  & 23.94 & 0.768 \\
 \hline
\end{tabular}
\caption{We show how the pool size used at test time in the FSA layer affects PSNR. Here we choose the pooling kernel size and stride to be equal to $p$. As expected, PSNR increases as pool size decreases. All models were trained for $3\times 10^4$ steps on 8 GPUs, using 11K large-scale Expedia Group lodging images and a crop size of 96px.}
\label{table:pooling}
\end{center}
\end{table}

\subsection{Visual Comparison}
Using the fine-tuned SRGAN$_{MSE}$, SRResNet, and A-SRResNet  described in the preceding sections, we generate SR images from 47 Google images in the public domain, and informally examine the results. Figure 6 shows four images from this set. We observe that in general, even when the fine-tuned model generates higher PSNR values, both end-to-end models tend to generate output with higher visual quality and fewer undesirable artifacts. Indeed, higher SSIM appears to correlate better with visual quality.

\begin{figure*}[h]
    \label{res}
        \begin{center}       
        \includegraphics[scale=0.28]{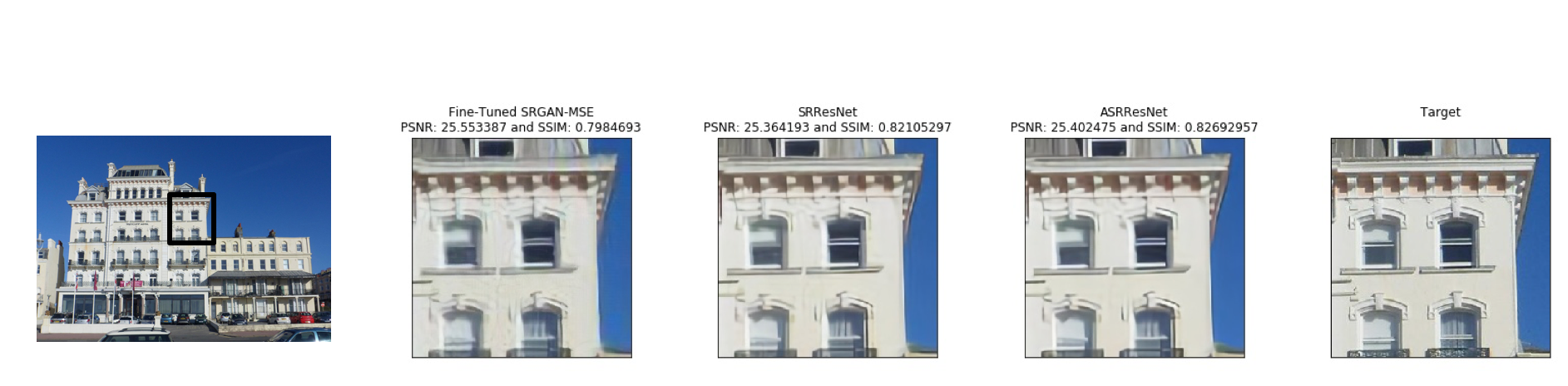}\\
        \includegraphics[scale=0.28]{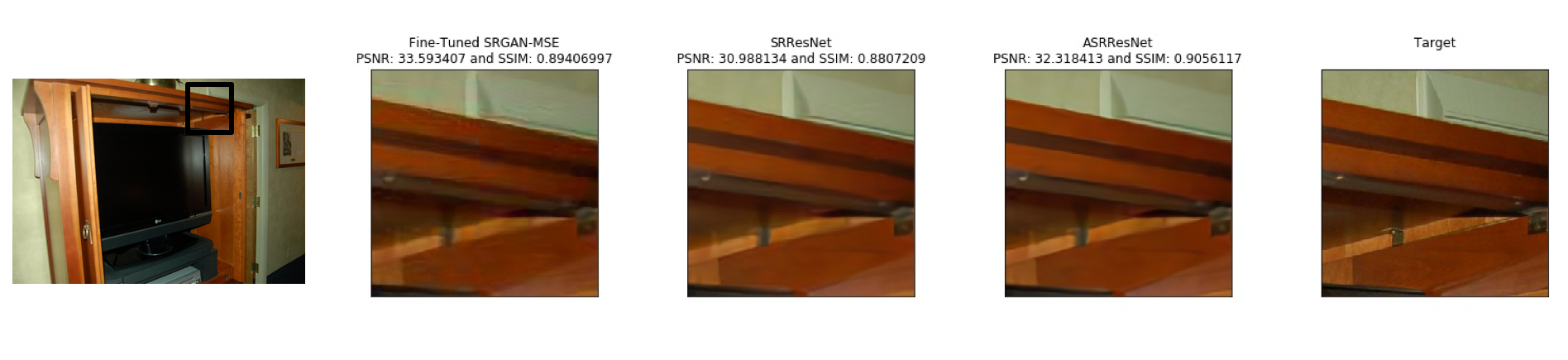}\\
        \includegraphics[scale=0.28]{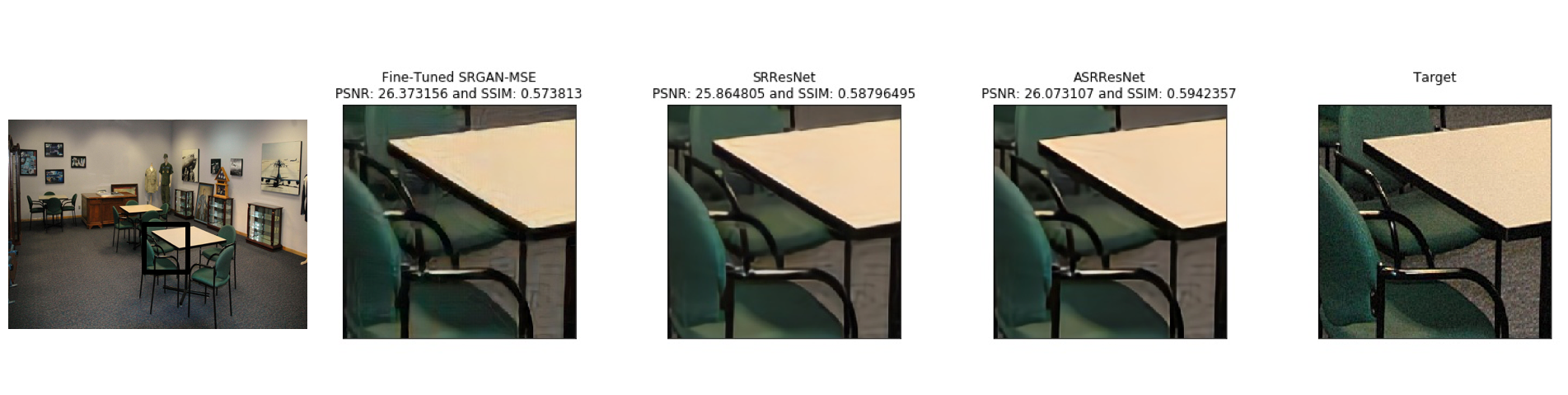}\\
        \includegraphics[scale=0.28]{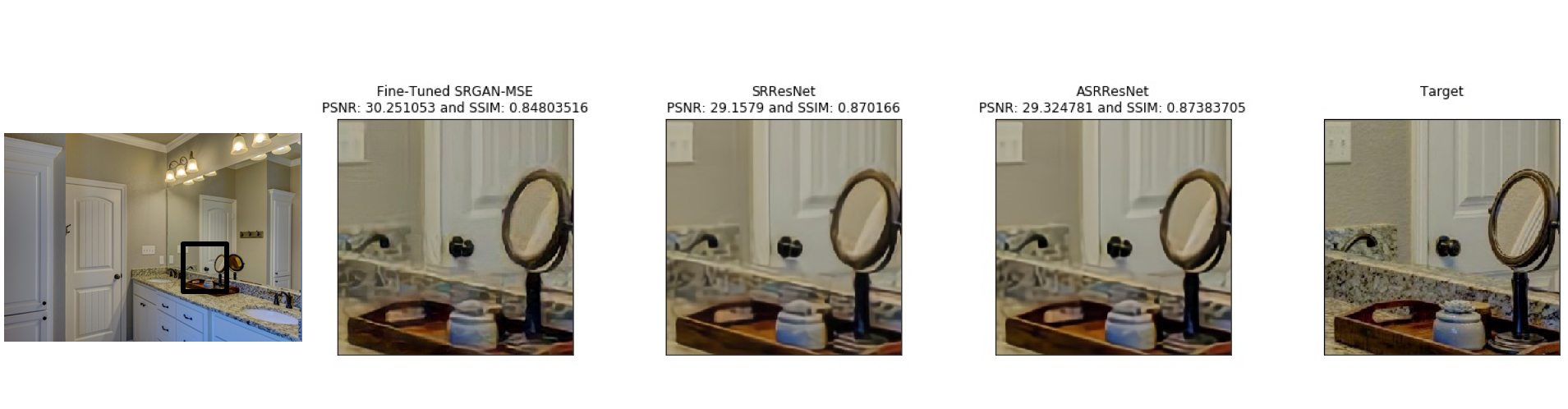}
        \caption{Four sample images from a dataset of high-resolution images from Google. The first column shows the original HR image. The remaining columns show the sub-image (black box in column 1) extracted from the output of three models (fine-tuned, SRResnet, and A-SRResNet) and the target image. In general, the end-to-end trained models produce noticeably superior results than the fine-tuned model, even when PSNR is higher with fine-tuning.}
        \end{center}
\end{figure*}

\section{Conclusion}
In this paper, we investigate several approaches for generating large-scale SR images with high visual quality: (1) application of a pre-trained SRGAN  classifier \cite{bworld}, (2) fine-tuning, and (3) training from scratch. We find that training A-SRGAN --- a novel attentional GAN model inspired by SRGAN \cite{DBLP:journals/corr/LedigTHCATTWS16}, SAGAN \cite{sagan}, and SNGAN \cite{snorm} --- in an end-to-end fashion using a large-scale lodging image data set generates high resolution images with higher SSIM score than other methods.  Both SRGAN and A-SRGAN handle texture quality and object consistency in the larger resolution space and can be used effectively with large-scale data. A-SRGAN adds a Flexible Self-Attention layer inspired by SAGAN \cite{sagan} to SRGAN, with pooling to reduce the size of the feature maps and enable large-size images to fit in memory. We also demonstrate that crop size is an important hyperparameter when working with large-scale images. Lastly, we describe a distributed algorithm which speeds up the training process for all of our models by almost a factor of five.

There are several directions for future work which could potentially improve the model. For one, we would like to investigate the impact of the location of attention layer within the network. Given that the first few layers of the network contain more pixel-level information than the later ones, we believe moving the attention layer to earlier layers could help model performance. Another important direction for future work is to enhance the model by first pre-training it on a very large set of images (such as ImageNet) and then fine-tuning it on our lodging dataset. In this way it would be easier for the model to reconstruct a variety of image textures. Also, to address the problem of how to choose the proper crop size of this model,  we would like to train an ensemble of models, each using a different crop size, and then aggregate the super-resolved images of these models to get the final output.

\section*{Acknowledgment}
This research is fully funded by Expedia Group. We would like to thank Expedia Group's Content Systems team for providing the HR dataset, and the EWE Data Science team for useful suggestions at all stages of this work.

\end{document}